\documentclass[nohyperref]{article}

\usepackage{microtype}
\usepackage{graphicx}
\usepackage{booktabs} % for professional tables

\usepackage{hyperref}

\usepackage[accepted]{icml2022}

\usepackage{amsmath}
\usepackage{amssymb}
\usepackage{mathtools}
\usepackage{amsthm}
\usepackage{xspace}
\usepackage{multirow}
\usepackage[table]{xcolor}% http://ctan.org/pkg/xcolor
\usepackage{xfrac}
\usepackage{graphicx}
\usepackage{caption}
\usepackage{subcaption}

\makeatletter
\newcommand{\thickhline}{%
    \noalign {\ifnum 0=`}\fi \hrule height 1pt
    \futurelet \reserved@a \@xhline
}
\makeatother

\usepackage[capitalize,noabbrev]{cleveref}

\definecolor{todo_color}{rgb}{1.0,0,0.0}
\definecolor{help_color}{rgb}{0.7,0.6,0.2}
\newcommand*{\erdosrenyi}{Erd\H{o}s-R\'enyi}

\definecolor{todo_color}{rgb}{1.0,0,0.0}
\definecolor{help_color}{rgb}{0.7,0.6,0.2}

\definecolor{mathias_color}{rgb}{.6,.4,.05}
\definecolor{edited_color}{rgb}{.5,.7,.1}
\definecolor{chengcheng_color}{rgb}{0.35,0,0}
\definecolor{chris_color}{rgb}{0,0.35,0}
\definecolor{cem_color}{rgb}{0,0,0.85}
\definecolor{markus_color}{rgb}{0.85,0.35,0.35}
\definecolor{rob_color}{rgb}{0.35,0.35,0}

\newcommand{\onedot}{\ifx\let\token.\else.\null\fi\xspace}
\newcommand{\ie}{i.e\onedot}
\newcommand{\eg}{e.g\onedot}

\newcommand{\ours}{GGR\xspace}

\newcommand{\first}[1]{\cellcolor{cyan!30}#1}
\newcommand{\second}[1]{\cellcolor{cyan!17}#1}
\newcommand{\third}[1]{\cellcolor{cyan!6}#1}

\newcommand{\firstB}[1]{\cellcolor{orange!30}#1}
\newcommand{\secondB}[1]{\cellcolor{orange!17}#1}
\newcommand{\thirdB}[1]{\cellcolor{orange!6}#1}

\theoremstyle{plain}

\theoremstyle{definition}

\theoremstyle{remark}

\icmltitlerunning{Gradient-based Weight Density Balancing in Dynamic Sparse Training}

\begin{document}

\twocolumn[
\icmltitle{Gradient-based Weight Density Balancing for Robust Dynamic Sparse Training}

% It is OKAY to include author information, even for blind
% submissions: the style file will automatically remove it for you
% unless you've provided the [accepted] option to the icml2022
% package.

% List of affiliations: The first argument should be a (short)
% identifier you will use later to specify author affiliations
% Academic affiliations should list Department, University, City, Region, Country
% Industry affiliations should list Company, City, Region, Country

% You can specify symbols, otherwise they are numbered in order.
% Ideally, you should not use this facility. Affiliations will be numbered
% in order of appearance and this is the preferred way.
\icmlsetsymbol{equal}{*}

\begin{icmlauthorlist}
\icmlauthor{Mathias Parger}{yyy}
\icmlauthor{Alexander Ertl}{yyy}
\icmlauthor{Paul Eibensteiner}{yyy}
\icmlauthor{Joerg H. Mueller}{yyy,huawei}
\icmlauthor{Martin Winter}{yyy,huawei}
\icmlauthor{Markus Steinberger}{yyy,huawei}
% \icmlauthor{Firstname7 Lastname7}{comp}
%\icmlauthor{}{sch}
% \icmlauthor{Firstname8 Lastname8}{sch}
% \icmlauthor{Firstname8 Lastname8}{yyy,comp}
%\icmlauthor{}{sch}
%\icmlauthor{}{sch}
\end{icmlauthorlist}

\icmlaffiliation{yyy}{Graz University of Technology, Austria}
\icmlaffiliation{huawei}{Huawei Technologies}
% \icmlaffiliation{sch}{School of ZZZ, Institute of WWW, Location, Country}

\icmlcorrespondingauthor{Mathias Parger}{mathias.parger@icg.tugraz.at}
\icmlcorrespondingauthor{Alexander Ertl}{ertl@student.tugraz.at}
\icmlcorrespondingauthor{Paul Einbensteinger}{eibensteiner@student.tugraz.at}
\icmlcorrespondingauthor{Joerg H. Mueller}{joerg.mueller@tugraz.at}
\icmlcorrespondingauthor{Martin Winter}{martin.winter@icg.tugraz.at}
\icmlcorrespondingauthor{Markus Steinberger}{steinberger@icg.tugraz.at}

% You may provide any keywords that you
% find helpful for describing your paper; these are used to populate
% the "keywords" metadata in the PDF but will not be shown in the document
\icmlkeywords{Machine Learning, ICML}

\vskip 0.3in
]
\thispagestyle{plain}
\pagestyle{plain}

% this must go after the closing bracket ] following \twocolumn[ ...

% This command actually creates the footnote in the first column
% listing the affiliations and the copyright notice.
% The command takes one argument, which is text to display at the start of the footnote.
% The \icmlEqualContribution command is standard text for equal contribution.
% Remove it (just {}) if you do not need this facility.

\printAffiliationsAndNotice{}  % leave blank if no need to mention equal contribution
% \printAffiliationsAndNotice{\icmlEqualContribution} % otherwise use the standard text.

\begin{abstract}
Training a sparse neural network from scratch requires optimizing connections at the same time as the weights themselves.
Typically, the weights are redistributed after a predefined number of weight updates, removing a fraction of the parameters of each layer and inserting them at different locations in the same layers.
The density of each layer is determined using heuristics, often purely based on the size of the parameter tensor.
While the connections per layer are optimized multiple times during training, the density of each layer remains constant.
This leaves great unrealized potential, especially in scenarios with a high sparsity of 90\% and more. 
We propose Global Gradient-based Redistribution, a technique which distributes weights across all layers - adding more weights to the layers that need them most.
Our evaluation shows that our approach is less prone to unbalanced weight distribution at initialization than previous work and that it is able to find better performing sparse subnetworks at very high sparsity levels. 
\end{abstract}

\section{Introduction}
With the rise of neural networks, plenty of research was conducted in optimizing network architectures \cite{Sandler2018,He2016}, training procedures \cite{Cubuk2019,Loshchilov2017} and inference hardware \cite{Chen2015,Han2016_EIE} to lower inference latency.
These achievements enabled the use of neural networks on low-power mobile devices like mobile phones or virtual reality headsets.
Yet, neural network inference remains computationally expensive, and the cost of training can be large - often taking days or weeks even in distributed environments.

In past years, quantization has established itself as a standard approach to reduce the memory footprint of the parameters and to accelerate inference as well as training using lower precision multiplication.
Another popular, yet not equally established, way to cut costs, is pruning, \ie, weight sparsification \cite{Le1990,Molchanov2017,Zhu2018}.
Like quantization, pruning lowers inference cost with little to no reduction in network performance.
The typical process of pruning consists of training the network densely and then iteratively removing weights which have the smallest impact on the accuracy of the prediction, while retraining the remaining weights.
This way, the number of weights is reduced significantly, reducing FLOPs and memory footprint at the same time.
Using smaller, dense networks from start does lower the training cost, but starting with a large model at first comes with the advantage of having a large set of potential subnetworks that can be established during early training.
Researchers have shown many times that large, sparse models outperform dense, small models with equal parameter count significantly \cite{Evci2019, Mostafa2019}.

With the requirements of a densely trained network and many iterations of pruning and parameter refinement, the training cost of pruned neural networks is typically much higher than for a dense neural network by itself.
Furthermore, memory consumption is often a limiting factor in the training process.
For example, the language model GPT-3 \cite{Brown2020} consists of 175 billion parameters - a multitude of what modern GPUs can store in memory.
Adding the size of feature maps, gradients and optimizer states, the memory consumption during training is much larger than in pure inference.
Recently, early pruning is becoming a popular research topic as it can reduce training cost compared to traditional pruning \cite{pmlr-v139-zhang21c,Chen2021,Liu2021_granet}.
These approaches start with a high density and then gradually reduce parameter density during training and before the network has converged.
Once parts of the weights start to settle, less important weights can be dropped iteratively.

Dynamic sparse training, on the other hand, starts with a sparse neural network from scratch and only requires a fraction of the parameters of the \emph{large}, dense model, and therefore also only a fraction of gradients, optimizer variables and FLOPs for processing them, while achieving comparable results.
% The lottery ticket hypothesis suggests that if a neural network is trained from scratch with only the sub-network that remains after pruning, the sparse network can be trained to the same accuracy as the original, dense model \cite{frankle2018the}.
Without knowing the set of connections upfront, however, a well-suited set of connections needs to be found at the same time as the weights are optimized.
Randomly connected networks fall short in accuracy compared to pruned models, especially at high sparsity levels.
Dynamic sparse training solves this by training the connections at the same time as the parameters.
This is accomplished using pruning-like concepts for weight removal, and special heuristics for weight insertion at set intervals during training.

Finding the best weights for removal and insertion is difficult and often either causes significant overhead which negates the gains of dynamic sparse training, or it fails to achieve results close to already highly optimized pruning techniques.
A common limitation of most approaches is that weights are only redistributed locally, on a per-layer basis.
The target sparsity of each layer is chosen heuristically at network initialization, for example by making larger layers more sparse and distributing the culled weights to smaller layers which likely need them more.
However, not all layers of a network require the same density, even if their parameter tensor has the same shape.
This problem becomes more apparent when using very high levels of sparsity ($>90$\%) where small differences in the density of a layer can make a big difference on the accuracy of the network.

We propose \emph{Global Gradient-based Redistribution} (\ours), a method which redistributes weights throughout layers dynamically during the training process by inserting the weights in the layers with the largest gradients.
Our main contributions are:
\begin{itemize}
    \item We propose a global gradient-based weight redistribution technique that adjusts layer densities dynamically during training.
    % \item We describe a memory-efficient strategy to implement our proposed weight redistribution strategy that overcomes the necessity of keeping all gradients in memory at once.
    \item We combine existing dynamic sparse training approaches to achieve the most robust and consistent method across a variety of configurations.
    \item We present ablation studies on various global weight redistribution schemes, comparing our proposed approach to current state-of-the-art metrics.
\end{itemize}
\section{Related Work}
Researchers have proposed many ways to reduce overparameterization of neural networks.
Typically, a network is first trained densely until convergence before starting an iterative pruning process.
Pruning removes weights with the lowest impact on the loss (\eg, weights with the smallest magnitude) and the remaining weights are shortly retrained to compensate for the loss of parameters \cite{Han2016,Molchanov2017,Zhu2018}.
Training a small, dense network with less neurons instead, however, does not result in the same accuracy as pruning a converged network.

The lottery ticket hypothesis suggests that when a working subnetwork is found (\eg, with pruning), the same subnetwork can achieve the same accuracy when trained from scratch \cite{frankle2018the}.
According to the authors of the lottery ticket hypothesis, the advantage of overparameterization is that during training, the network can find working subnetworks among large amounts of possible candidates.
Training a found subnetwork, the \emph{winning ticket}, to achieve the same accuracy, however, is difficult and often requires initializing the subnetwork with weights extracted from a dense network trained for thousands of optimizer steps \cite{Zhou2019}.
However, the hypothesis shows that at least simpler architectures can be trained with a sparse neural network from scratch.
The challenge remains to find a combination of connections that achieves a high accuracy - without having to train the network densely first to extract the winning ticket.

The concept of training a randomly initialized sparse neural networks from scratch was introduced in 2018 with two different approaches.
\emph{Deep Rewiring} (DeepR) \cite{Bellec2017} proposes to drop weights that would flip the sign during the optimizer update step.
The same number of weights is then inserted randomly in the same layer and initialized with a random sign.
\emph{Sparse Evolutionary Training} (SET) \cite{Mocanu2017}, on the other hand, makes use of a simple and well-established weight removal approach in pruning research.
Following the intuition that weights with a smaller magnitude contribute less to the output of a neuron, SET drops a fixed number of weights with the lowest magnitude in predefined intervals and inserts the same amount randomly.
RigL \cite{Evci2019} improved the criteria introduced in SET.
While they also remove the weights with the lowest magnitude, they insert the set of weights with the highest gradients instead of selecting them randomly.
This way, connections that could have the greatest impact on the result of the current mini-batch are inserted first.

The common limitation of all three approaches is that they use a predefined density per layer.
This does have the advantage that the network can be designed to reach a FLOP target in inference. 
However, with deep neural networks, selecting the right density per layer is difficult and performance is likely left on the table.

\emph{Dynamic sparse reparameterization} (DSR) \cite{Mostafa2019} proposes dynamic weight distribution across layers to mitigate this issue.
They drop connections with a magnitude below a threshold which is dynamically adjusted to get close to a target drop rate.
The removed weights are then distributed among the sparse weight tensors following the heuristic that layers with a larger number of remaining weights should receive a larger fraction of the available weights.
The intuition behind it is that layers which can drop more weights might not need as many weights as denser layers which could not afford to lose weights in the first place.
This number of weights is then placed randomly at zero-valued positions in the respective layers.
One downside of their approach is that layers that are initialized very sparsely will take a long time to reach their optimum density level, because they receive a smaller fraction of the available weights.
This problem becomes more visible with very high sparsity levels ($>90\%$) where some layers contain only a few dozen or less parameters.
% \emph{Sparse networks from scratch} (SNFS) \cite{Dettmers2019} proposes to distribute weights across layers based on their momentum.
% This approach requires to process dense gradients for all parameters at every training step and to store the dense momentum of each parameter at all times. 
% Thus, it mitigates most of the benefits of sparse learning.
% However, in contrast to traditional pruning, it is still considered a dynamic sparse training approach as it always uses a sparse network in the forward propagation step. \todo{this paper is only on arxiv. don't know if we want to compare against them. it was rejected from ICLR}

We combine the approaches of RigL and DSR, allowing for dynamic weight distribution across layers, but using the gradients as criteria instead of the layer density.
We distribute weights across layers using the top K largest gradients during the redistribution step - information that is already available when using RigL for weight insertion.
Using gradients instead of density as a criterion makes our approach more robust against the initial distribution.
Very sparse layers that require more weights can get them quickly, potentially even become fully dense after a single weight redistribution iteration if this layer is determined as the main bottleneck.
% This way, we only need to compute dense gradients during the redistribution step and we have a much lower memory overhead, processing one layer at a time while keeping the top K gradients in memory.
% In contrast to SNFS, however, we found that it is not required to use the same criteria for selecting the number of weights inserted by layer and selecting the specific weights to be inserted per layer.
We found that weight distribution across layers and weight insertion inside a layer can benefit from using different criteria.
A combination of gradient-based distribution of a number of weights to be inserted per layer and a random insertion per layer can potentially improve the overall performance compared to purely gradient-based approaches.

% \todo{cite  in-time dense overparameterization \cite{Liu2021} }

Utilizing sparsity in the neural network for accelerating inference and training, however, is non-trivial.
Dense matrix-matrix multiplication is one of the best use-cases of modern GPUs and special neural accelerator hardware.
Sparse matrices are much more complex to process on SIMD devices, requiring load balancing or matrix analysis to fully utilize GPUs.
While some solutions were proposed for achieving real-world speedups on modern inference hardware \cite{Huang2019,Wang2020,Zhou2021,Liu2021_inference,chen2022pixelated}, we only focus on parameter density vs. prediction accuracy as done in previous work.

\section{Method}
\begin{figure*}[ptb]
    \centering
    \includegraphics[width=0.9\textwidth]{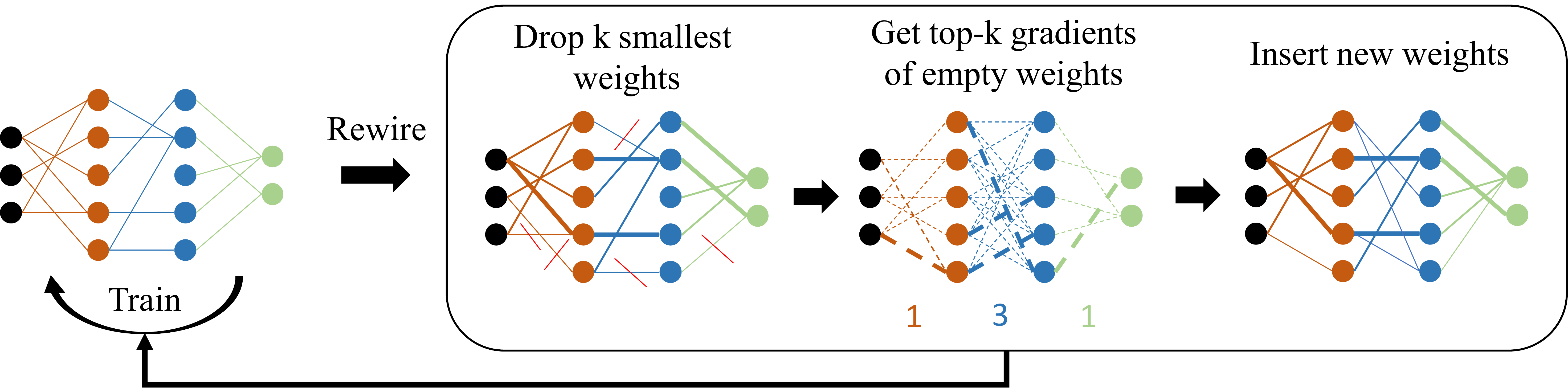}
    \caption{Training routine of \ours. After training the network for a given number of mini-batches, the network is rewired. \ours uses magnitude pruning to drop a given ratio of smallest weights per layer. The same number of weights is then distributed across the network. We determine the number of weights to be inserted per layer by extracting the top-k gradients of all zero-valued weights (1,3,1 in this example). After that, this number of weights is inserted into the respective layers using a combination of gradient-based (RigL) and random (SET) insertion.}
    \label{fig:ggr_training_routine}
\end{figure*}

% \markus{I think the method section in general is too short. likely needs some figures and more insights}
% The main steps of the training process are the same as in previous work on dynamic sparse training.
Dynamic sparse training starts by initializing the network randomly using a target number of parameters per layer.
After training the network for a given number of optimizer steps, a predefined ratio of least significant weights is removed and inserted where we expect the weights to contribute more to a reduction of the training loss.
These steps are repeated for a given number of rounds, lowering the ratio of weights to distribute with each step.
Then, the final architecture is trained for the remaining epochs without weight redistribution.
The specific implementation details of our approach are described below and visualized in Figure~\ref{fig:ggr_training_routine}.

\subsection{Network initialization}
Selecting a good initial weight distribution is a difficult process that can have a great impact on the accuracy of the network.
With our dynamic weight redistribution technique, however, we are less sensitive to network initialization.
We implement two different approaches for selecting the layer density at initialization: \erdosrenyi-Kernel (ERK) \cite{Mocanu2018,Evci2019} and uniform density.
The \erdosrenyi-Kernel scales the density with the number of inputs and outputs of each layer.
The density is scaled proportional to $1-\frac{n^{l-1}+n^l+w^l+h^l}{n^{l-1} \times n^l\times w^l\times h^l}$, where $n^l$ and $n^{l-1}$ is the number of outputs (neurons) and inputs of layer $l$, and $w^l$ and $h^l$ are the width and height of the convolutional kernel.
For linear layers, $w^l$ and $h^l$ are removed from the equation.
With the \erdosrenyi-Kernel, smaller layers get a greater share of the available weights than large layers.
This heuristic works well as a first approximation and is often used in approaches without global weight redistribution.
With uniform density, we use the same density across all layers, including the first and last layer.
While it is well known that much higher performance can be achieved using a higher density for those layers, this serves as a great benchmark for evaluating the density distribution capabilities of \ours versus existing density adjusting approaches.

\subsection{Weight removal}
After training the weights for a few mini-batches, the least important weights are removed.
The percentage of weights to be redistributed is decreased every epoch using the cosine-function as suggested by the authors of RigL \cite{Evci2019}.
Performing redistribution steps too often could increase the risk of mainly deleting weights that were just inserted.
We mitigate this risk by training the weights for a predefined number of optimizer steps to let the newly inserted weights integrate into the existing network.
After that, we remove the weights with the smallest magnitude (magnitude pruning).
These smallest weights can either be selected locally for each layer using a uniform percentage of active weights to be removed, or globally for the entire network.
Early evaluation showed a minor advantage of local weight removal over the global counterpart.
Because of that, we decided to only focus on local weight removal as is the standard in previous work.

\subsection{Global weight redistribution}
\label{sec:global_weight_redistribution}
After deleting the least significant weights, the same number of weights is inserted back into the network.
We split this part up into two independent steps: 1) determine the number of weights to be inserted per layer and 2) select specific weights to insert.
% For step 1), we sort the gradients of all zero-valued (non-existing) weights of all layers by magnitude and crop the sorted list to the number of removed weights.\markus{I dont understand:}
% The number of weights to be inserted per layer is determined by counting the number of remaining weights per layer in a sorted array. 
For step 1), we extract the top-$k$ gradient magnitudes of all layers, where $k$ is the number of weights removed in the previous step.
The number of top-$k$ gradients belonging to a layer $l$ determines the number of new weights to be inserted into layer $l$.
Extracting the top-k gradient magnitudes does not require keeping all gradients in memory at once, nor to perform global sorting.
We reduce the memory consumption significantly by iteratively comparing the current top-k array vs the gradients of one layer.
By only taking the top-k values of the union of the current top-k gradients and the gradients of this layer, we achieve the same result with a lower peak memory consumption than with global sorting.

In step 2), one could use various existing approaches to insert the number of weights determined per layer.
% In this work, we evaluate three different weight insert criteria: we either insert weights with the largest gradient (RigL), we insert them randomly (SET) or we combine both with a 50:50 split.
% We compare all three approaches in our evaluation.
Early experiments showed that existing weight insertion techniques achieve very inconsistent results across different datasets, densities and network architectures.
Gradient-based insertion (RigL) can outperform random insertion (SET) significantly in some experiments while falling back in others.
We found that combining both approaches makes training more robust, achieving more consistent performance across scenarios.
Thus, we decided to use a 50:50 split, \ie, we insert 50\% of the weights per layer using the largest gradients and 50\% randomly across the parameter tensor. 
Previous work shows that zero-initialization of newly inserted weights leads to equal or even slightly better performance than gradient-based insertion \cite{Evci2019}.
Because of that, we initialize newly inserted weights with a small, but non-zero value of $1^{-10}$ as zero-valued weights would be pruned away on insertion.

% \markus{what about the contribution point on memory efficient implementation?}
\section{Evaluation}
We evaluate \ours on datasets of different complexity and networks of different size.
% In our evaluations, we want to answer the question how the commonly used \erdosrenyi kernel distribution compares to dynamic distribution.
% We compare our dynamic global density distribution method against DSR \cite{Mostafa2019} and RigL \cite{Evci2019}, which uses the same criteria for weight distribution, but with a fixed parameter count per layer.
% We evaluate these approaches on datasets of different complexity and networks of different size.
% As small network baseline, we chose to use MLP-Mixer \cite{Tolstikhin2021}, a MLP architecture designed for computer vision tasks.
% MLP-Mixer comes with many configurations and scales well from small to large problems.
% We chose to use a small custom configuration (patch resolution=4x4, hidden size=256, $D_{C}$=512, $D_{S}$=128, layers=8) with \todo{how many?} weights.
As a small network baseline, we use MobileNetv2 \cite{Sandler2018} with 7.4M parameters.
MobileNetv2 is a FLOP-efficient CNN architecture that achieves a good trade-off between inference cost and prediction accuracy.
As a medium-sized network, we use ResNet-18 \cite{He2016} with 22.4M weights.
As a large-sized baseline, we use the expensive VGG16 network \cite{Simonyan2015} with 138M parameters.
VGG16 is not a state-of-the-art, parameter-efficient network nowadays, but it is a good showcase for a waste of parameters when used on small datasets.
The first fully connected layer, for example, consists of over 102 million weights out of 138 million parameters used in total by the network.
If \ours's weight redistribution criteria perform well, this layer should become very sparse, while the convolutional layers get a larger share of the weights.

We train the models on CIFAR-100 \cite{Krizhevsky09learningmultiple} and Tiny-ImageNet \cite{Le2015} with 100 and 200 classes respectively.
With the different dataset complexities, we can compare how well our approach works when the network is large enough to afford losing some parameters and when the network is limited by its capacity even in the dense case.
% We train the network on CIFAR-100 and Tiny-ImageNet, with input resolutions of 32x32 and 64x64 pixels respectively.
VGG16 is designed for a higher resolution input than those in our datasets.
For our evaluations on the Tiny-ImageNet dataset with only 64x64 pixel inputs, we adapted the network.
We remove the last block of convolutions and maximum pooling layers before the linear layers - reducing the downscale of the input image while removing the most expensive convolutional layers. 
With this adjustment, VGG16 uses 58.8M parameters for Tiny-ImageNet instead of 138M in its original design.

All configurations are trained using randomly initialized weights - the connectivity is selected as a uniform random pattern, and the weights are initialized using uniform Kaiming initialization \cite{He2015_kaiming}.
We train the network over $200$ epochs using a batch size of 128 and stochastic gradient descent with a learning rate of $0.1$ and a downscale of factor $5$ at epoch $60$, $120$ and $160$.
The first epoch is used as a warmup epoch, gradually increasing the learning rate from $0.0$ to $0.1$.
The ratio of weights redistributed at the beginning of every epoch, except the first, is reduced every epoch using the cosine function starting with $10\%$ at epoch $1$ and $0\%$ at epoch $100$.
The final selection of connections is then trained for the remaining $100$ epochs to allow the network to converge.
In the case of CIFAR-100, we found that we achieve the same results with only $100$ epochs instead of $200$.
All above mentioned epochs are divided by $2$ respectively.

% \todo{should we use them in the end? }
% We use the augmentations proposed by \emph{Autoaugment} \cite{Cubuk2019} for the CIFAR datasets.
% For Tiny-ImageNet, we use the ImageNet augmentations. 
\section{Results}

\begin{table*}[ptb]
\centering
\setlength{\tabcolsep}{4pt}
\renewcommand\arraystretch{1.0}
\resizebox{0.495\linewidth}{!}{
\begin{tabular}{c|c||cc|cc|cc}
\hline\thickhline
\multirow{2}{*}{Network} & \multirow{2}{*}{Method} & \multicolumn{2}{c|}{s=0.9} & \multicolumn{2}{c|}{s=0.97} & \multicolumn{2}{c}{s=0.99} \\
\cline{3-8}
& & ERK & uni & ERK & uni & ERK & uni \\
\hline
\multirow{5}{*}{MobileNetv2} 
& Dense & \multicolumn{6}{c}{69.15} \\
& SET       & 66.34         & 55.31         & 56.13         & - & 25.75         & -  \\
& RigL      & \second{67.02}& \third{58.13} & \third{58.96} & - & \second{43.79}& - \\
% & SET+RigL  & 66.52         & 54.28         & 55.77         & - & 39.64         & -  \\
& DSR       & \first{68.26} & \first{65.85} & \first{61.11} & - & \third{37.38} & - \\
& GGR      & \third{66.79} & \second{64.55}& \second{60.57}& - & \first{46.49} & - \\
\hline
\multirow{5}{*}{ResNet-18} 
& Dense & \multicolumn{6}{c}{76.27} \\
& SET       & \first{74.18} & 71.64         & 70.13         & 66.64         & 63.45         & 57.78 \\
& RigL      & \second{74.03}& \third{72.15} & \second{70.65}& \third{68.31} & \first{65.48} & \third{60.81} \\
% & SET+RigL  & 74.06         & 72.09         & 70.74         & 67.87         & 64.21         & 59.21  \\
& DSR       & \third{73.89} & \first{73.47} & \first{70.99} & \second{70.17}& \third{64.12} & \second{63.77} \\
& GGR      & 73.34         & \second{72.83}& \third{70.28} & \first{70.71} & \second{65.12}& \first{64.41} \\
\hline\thickhline
\multicolumn{8}{c}{\multirow{2}{*}{CIFAR-100}} \\
\end{tabular}
}
% \caption{
% Top-1 accuracy on the test set of CIFAR-100. Every method is trained with 90\%, 97\% and 99\% sparsity ($s$) using uniform and \erdosrenyi-Kernel density distribution at initialization.
% Ours G, Ours R and Ours G+R use gradient based (G), random (R) and combined (G+R) local weight insertion per layer.
% Configurations with an accuracy below 5\% are not reported as these values do not qualify for comparisons.
% }
\label{tab:results_CIFAR100}
% \todo{fuse tables? they could share the caption at least}
%
\resizebox{0.495\linewidth}{!}{
\begin{tabular}{c|c||cc|cc|cc}
\hline\thickhline
\multirow{2}{*}{Network} & \multirow{2}{*}{Method} & \multicolumn{2}{c|}{s=0.9} & \multicolumn{2}{c|}{s=0.97} & \multicolumn{2}{c}{s=0.99} \\
\cline{3-8}
& & ERK & uni & ERK & uni & ERK & uni \\
\hline
\multirow{5}{*}{MobileNetv2} 
& Dense & \multicolumn{6}{c}{55.77} \\
& SET       & 53.35         & \third{47.73} & 43.00         & - & - & - \\
& RigL      & \second{55.08}& \third{49.85} & \first{48.23} & - & - & - \\
% & Set+RigL  & 55.34         & 49.75         & 46.53         & - & - & - \\
& DSR       & \third{54.48} & 38.80         & \third{45.06} & - & - & - \\
& GGR      & \first{55.31} & \second{52.43}& \second{48.16}& - & - & - \\
\hline
\multirow{5}{*}{VGG16} 
& Dense & \multicolumn{6}{c}{60.88} \\
& SET       & \first{61.04} & 57.44         & \third{57.04} & \third{52.40} & \second{53.14}& \second{45.34} \\
& RigL      & \second{60.31}& \third{58.31} & 55.82         & \third{52.40} & 51.28         & -     \\
% & SET+RigL  & 59.94         & 58.44         & 55.87         & 52.14         & 50.99         & -  \\
& DSR       & \third{59.94} & \first{60.04} & \second{57.13}& \second{56.42}& \first{53.47} & -  \\
& GGR      & 59.04         & \second{58.65}& \first{57.60} & \first{56.95} & \third{52.89} & \first{52.30} \\

\hline\thickhline
\multicolumn{8}{c}{\multirow{2}{*}{Tiny-ImageNet}} \\
\end{tabular}
}
\caption{
Top-1 accuracy in percent on the test set of CIFAR-100 and Tiny-ImageNet. Every method is trained with 90\%, 97\% and 99\% sparsity ($s$) using uniform (uni) and \erdosrenyi-Kernel (ERK) density distribution at initialization.
% Ours G, Ours R and Ours G+R use gradient based (G), random (R) and combined (G+R) local weight insertion per layer.
Colored values highlight the three best performing approaches.
Configurations with an accuracy below 5\% are not reported as these values do not qualify for comparisons.
}
\label{tab:results_tinyimagenet_CIFAR100}
\end{table*}

\begin{table}[ptb]
\centering
\renewcommand\arraystretch{1.2}
\resizebox{\linewidth}{!}{
\begin{tabular}{c||cc|cc|cc}
\hline\thickhline
\multirow{2}{*}{Method} & \multicolumn{2}{c|}{s=0.9} & \multicolumn{2}{c|}{s=0.97} & \multicolumn{2}{c}{s=0.99}\\
\cline{2-7}
& ERK & uni & ERK & uni & ERK & uni \\
\hline
Dense & \multicolumn{6}{c}{65.52} \\
SET       & \third{63.73}   & 58.03         & 56.58         & 59.52         & 47.45         & \second{51.56}   \\
RigL      & \second{64.11}  & \third{59.61} & \third{58.42} & \third{60.36} & \second{53.52}& -       \\
DSR       & \first{64.14}   & \second{59.64}& \second{58.57}& \second{63.30}& \third{51.66} & -       \\
GGR       & 63.62           & \first{62.11} & \first{59.15} & \first{63.83} & \first{54.83} & \first{58.36}   \\
\hline\thickhline
\end{tabular}
}
\caption{
Top-1 accuracies averaged over all models and datasets. When all methods failed in the same scenario, we report the average only over the set of working scenarios for a given sparsity. In the last column, we do not report the values for RigL and DSR since they were not able to fit VGG-16 on TinyImageNet in contrast to SET and \ours.
}
\label{tab:avg_results}
\end{table}

% \todo{rewrite section - focus more on average performance than on individual results.}
Our evaluation shows that \ours achieves state-of-the-art results in all configurations (see Table~\ref{tab:results_tinyimagenet_CIFAR100}) .
Highly sparse configurations with a sparsity of 97\% and 99\% benefit most from dynamic density adjustment.
The results suggest that our approach is less sensitive to unbalanced layer densities at initialization than previous work.
The best example for that is VGG16 which fails to learn using RigL and DSR using a uniform layer sparsity of 99\%, while \ours performs nearly as good as with ERK initialization.
At the same time, our approach, like all other approaches, is unable to to fit the model in case of very bad initial distribution.
MobileNetv2, with a sparsity at or above 97\% using a uniform density distribution, is not able to recover from initial distribution.
With uniform density, some small layers contain less than a dozen weights at initialization - breaking the gradients during backpropagation.
However, MobileNetv2 can be trained on the CIFAR-100 dataset with a 99\% weight sparsity using ERK.
\ours achieves by far the best results in this setting, yet also with a large reduction in accuracy compared to the 97\% sparsity configuration.

Looking at the top-1 accuracies achieved on average with each method for a specific sparsity and initialization (Table~\ref{tab:avg_results}), we can see that \ours achieves the best results with high sparsity or bad initial density.
This suggests that \ours is better at distributing weights to layers that need them most compared to DSR.
At the same time, DSR achieves best results with 90\% sparsity as long as \erdosrenyi-Kernel is used for initialization.
Looking at the individual results with a sparsity of 90\% and ERK reveals that \ours is performing worse than previous work when applied to large models.
To our surprise, SET achieves the best performance both with ResNet-18 and VGG16.
In the case of VGG16, it is even able to outperform the dense reference, which suggests that random insertion is able to reduce overfitting which inevitably happens when using such a large network for this relatively small dataset.
To confirm this, we trained \ours only using either gradient-based insertion or random insertion using a sparsity of 90\% and ERK.
Gradient-based insertion performs significantly worse, resulting in an accuracy of only 57.70\%, while random insertion achieves a much better accuracy with 60.42\%.

% To our surprise, the three different weight insertion techniques (gradient-based, random, half gradient-based/half random) perform very inconsistently across different architectures and datasets.
% We noticed during early evaluations that gradient-based and random insertion can each outperform the other approach significantly in different scenarios.
% Our intuition was that a combination of gradient-based and random insertion might result in a more consistent and potentially overall favorable performance.
% While this proved to be true in most cases, the combined insertion technique can also perform worse than either, as can be seen in the 90\% sparse configuration when training ResNet-18.

VGG16 is the largest model in our evaluation and therefore has the greatest potential for sparsification.
With a sparsity of 90\%, all approaches are able to achieve an accuracy close to the dense reference while SET even exceeds the accuracy of the dense model.
Surprisingly, \ours outperforms the dense MobileNetv2 by 1.83\% using only 3\% of VGG16's weights.
In other words, VGG16 with 1.76M parameters can outperform the dense and efficient MobileNetv2 with 7.4M parameters.
We see the same behavior with ResNet-18 and MobileNetv2.
With only 3\% of the parameters of the dense ResNet-18 model, all sparse redistribution methods are able to outperform dense MobileNetv2.
For each frame in CIFAR-100, ResNet-18, trained with \ours using 97\% sparsity, processes 80 MFLOPs during inference compared to 119 MFLOPs with the dense MobileNetv2 - at 1.13\% higher top-1 accuracy.

\subsection{Additional Evaluations}
% \begin{table}[ptb]
% \centering
% \setlength{\tabcolsep}{4pt}
% \renewcommand\arraystretch{1.2}
% \resizebox{\linewidth}{!}{
% \begin{tabular}{c|c||cc|cc|cc}
% \hline\thickhline
% \multirow{2}{*}{Network} & \multirow{2}{*}{Method} & \multicolumn{2}{c|}{Empty Neurons} & \multicolumn{2}{c|}{Empty Inputs} & \multicolumn{2}{c}{Density Variance} \\
% \cline{3-8}
% & & ERK & uni & ERK & uni & ERK & uni \\
% \hline
% \multirow{5}{*}{MobileNetv2} 
% & RigL      & 64.1\%    & -     & 67.6\%    & - & -     & - \\
% & DSR       & 59.8\%    & -     & 55.5\%    & - & -     & - \\
% & Ours G    & 76.2\%    & -     & 70.6\%    & - & -     & - \\
% & Ours R    & 51.1\%    & -     & 28.6\%    & - & -     & - \\
% & Ours G+R  & 64.6\%    & -     & 63.6\%    & - & -     & - \\
% \hline
% \multirow{5}{*}{VGG16}
% & RigL      & 53.0\%    & 59.8\%    & 34.0\%    & 53.6\%    & -     & - \\
% & DSR       & 56.6\%    & -         & 51.7\%    & -         & -     & - \\
% & Ours G    & 66.2\%    & 75.1\%    & 65.3\%    & 67.7\%    & -     & - \\
% & Ours R    & 23.7\%    & 22.1\%    & 10.5\%    & 17.6\%    & -     & - \\
% & Ours G+R  & 65.0\%    & 72.7\%    & 56.5\%    & 62.7\%    & -     & - \\
% \hline\thickhline
% \end{tabular}
% }
% \caption{
% Tiny-ImageNet at 97\% sparsity.
% }
% \label{tab:empty_neurons_inputs}
% \end{table}

\begin{table}[ptb]
\centering
\renewcommand\arraystretch{1.2}
\resizebox{\linewidth}{!}{
\begin{tabular}{c||cc|cc|cc}
\hline\thickhline
\multirow{2}{*}{Method} & \multicolumn{2}{c|}{s=0.9} & \multicolumn{2}{c|}{s=0.97} & \multicolumn{2}{c}{s=0.99}\\
\cline{2-7}
& ERK & uni & ERK & uni & ERK & uni \\
\hline
SET       & \thirdB{63.73} & \thirdB{58.03}   & \thirdB{56.58} & \thirdB{59.52}  & \thirdB{47.45} & \firstB{51.56}   \\
RigL      & \firstB{64.11}  & \firstB{59.61}  & \firstB{58.42} & \firstB{60.36}  & \firstB{53.52}  & -       \\
Set+Rigl  & \secondB{63.97} & \secondB{58.64} & \secondB{57.23}& \secondB{60.01} & \secondB{51.61} & - \\
\hline
GGR SET  & \first{64.07}   & \first{63.63} & \third{57.33} & \first{64.11} & \third{46.53}  & \second{58.70}   \\
GGR RigL & \third{62.21}   & \third{60.20} & \second{58.36}& \third{62.96}  & \second{53.03}  & \first{59.44}   \\
GGR     & \second{63.57}  & \second{62.11}& \first{59.15} & \second{63.83}  & \first{54.83}  & \third{58.36}   \\
\hline\thickhline
\end{tabular}
}
\caption{
Top-1 accuracies averaged over all models and datasets. When all methods failed one configuration, we report the average over those configurations that were able to fit the data. In the last column, we do not report the values for RigL and DSR since they were not able to fit VGG-16 on TinyImageNet in contrast to SET and \ours.
}
\label{tab:avg_results_ablation}
\end{table}

\begin{table}[ptb]
\centering
\renewcommand\arraystretch{1.2}
\resizebox{\linewidth}{!}{
\begin{tabular}{c|c||cc|cc|cc|cc}
\hline\thickhline
\multirow{2}{*}{Net} & \multirow{2}{*}{Method} & \multicolumn{2}{c|}{Top-1 Acc.} & \multicolumn{2}{c|}{Empty Neurons} & \multicolumn{2}{c|}{Empty Inputs} & \multicolumn{2}{c}{Params* [M]}\\
\cline{3-10}
& & ERK & uni & ERK & uni & ERK & uni & ERK & uni \\
\hline
\multirow{4}{*}{MNv2} 
& SET       & 43.00 & -     & 50.2  & 37.9  & 54.4  & 7.5 & 2.48 & 6.74  \\
& RigL      & 48.23 & -     & 68.6  & 38.1  & 73.7  & 7.7 & 0.89 & 6.74  \\
& DSR       & 45.06 & -     & 72.4  & 37.5  & 67.2  & 7.8 & 1.87 & 6.74   \\
% & Ours G    & 47.20 & -     & 81.8  & 38.4  & 76.1  & 7.4 & 0.80 & 6.74   \\
% & Ours R    & 44.32 & -     & 59.6  & 38.0  & 36.8  & 7.3  & 2.96 & 6.75  \\
& GGR  & 48.16 & -     & 71.5  & 38.0  & 70.5  & 7.3 & 1.22 & 6.74   \\
\hline
\multirow{4}{*}{VGG16}
& SET       & 57.04 & 52.40 & 31.9  & 37.8  & 41.5  & 43.6 & 17.4 & 17.9  \\
& RigL      & 55.82 & 52.40 & 49.6  & 61.7  & 42.2  & 53.6 & 3.20 & 3.27  \\
& DSR       & 57.13 & 56.42 & 57.1  & 68.2  & 54.1  & 64.2 & 3.95 & 1.91  \\
% & Ours G    & 56.91 & 56.04 & 62.6  & 73.5  & 69.3  & 63.6 & 1.18 & 2.52 \\
% & Ours R    & 58.57 & 58.43 & 29.7  & 30.2  & 35.5  & 34.6 & 20.6 & 20.4  \\
& GGR  & 57.60 & 56.95 & 61.8  & 70.2  & 63.9  & 56.2 & 1.44 & 3.33  \\
\hline\thickhline
\end{tabular}
}
\caption{
Percentage of empty neurons and inputs after training MobileNetv2 (MNv2) and VGG16 on Tiny-ImageNet at 97\% sparsity. For convolutional layers, we report the percentage of empty output channels and empty input channels.
* the number of parameters is the number of dense parameters that remain when the original parameter is scaled down in size by removing all completely empty neurons and inputs. 
Downsized parameter tensors are easier to accelerate on modern hardware than a sparse parameter tensor. However, none of these approaches target a high neuron or input sparsity by design.
}
\label{tab:empty_neurons_inputs}
\end{table}

As mentioned in Section~\ref{sec:global_weight_redistribution}, early experiments showed inconsistent performance when only relying on either gradient-based or random insertion.
We perform an ablation study comparing SET, RigL and SET \& RigL combined using local distribution as well as with \ours's global distribution (see Table~\ref{tab:avg_results_ablation}). 
This ablation study confirms our early experiments and shows that while using one redistribution method can outperform the combined approach in four out of six cases, their individual performance is much more inconsistent. 
They can fall behind each other by a large margin while the combined approach is always close to the best of each.
Interestingly, the same does not apply to local distribution where RigL always achieves the best results except for the right-most column (uniform layer sparsity of 99\%).

Utilizing sparsity on SIMD devices for faster inference is difficult with modern accelerator architectures built around dense data.
An easy-to-implement approach is to prune entire neurons as this leads to smaller parameters tensors and output feature maps, which works well together with existing, dense frameworks.
For better understanding, assume that half of the neurons do not use a single weight and half of the inputs are not used by a single neuron - in this case, we could lower the size of the parameter tensor to $\sfrac{1}{4}$ of the original size - making it much easier to be used to speedup inference on dense hardware accelerators.
Comparing the final subnetwork that evolved during training shows some interesting differences in how parameters are distributed across and inside of parameter tensors.
Table~\ref{tab:empty_neurons_inputs} shows how many of the neurons and inputs are used in the end when trained on Tiny-ImageNet with a sparsity of 97\%.
MobileNetv2 initialized with \erdosrenyi-Kernel results in similar neuron and input sparsities with all approaches except for SET which uses more neurons and inputs - while achieving the lowest performance.
In the case of VGG16, however, \ours achieves the best accuracy with the greatest neuron and input sparsity when using \erdosrenyi-Kernel initialization.
With uniform initialization, the final network is very different, and less sparse overall.
While \ours still achieves the best accuracy, the network cannot be shrunk as well as with ERK initialization.

Table~\ref{tab:empty_neurons_inputs} also gives insights into models that fail to fit the data (MobileNetv2 with uniform initialization).
All methods show similar counts of empty neurons and empty inputs. 
None of them are able to rewire the network successfully.
Compared to models that are able to fit the data, the ratio of empty neurons and inputs is surprisingly much lower.
Yet, the network is still unable to properly propagate inputs and gradients.

\begin{figure*}[ptb]
    \centering
    \includegraphics[width=\textwidth]{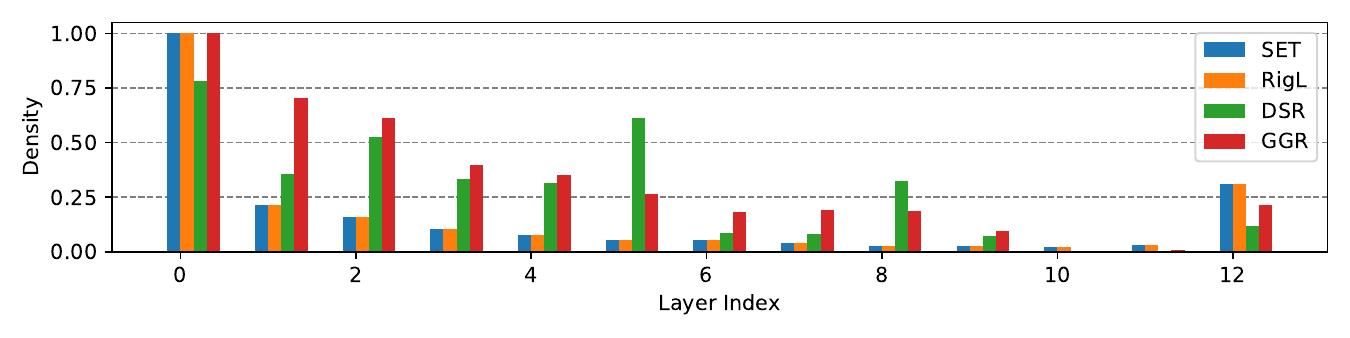}
    \vspace{-10pt}
    \includegraphics[width=\textwidth]{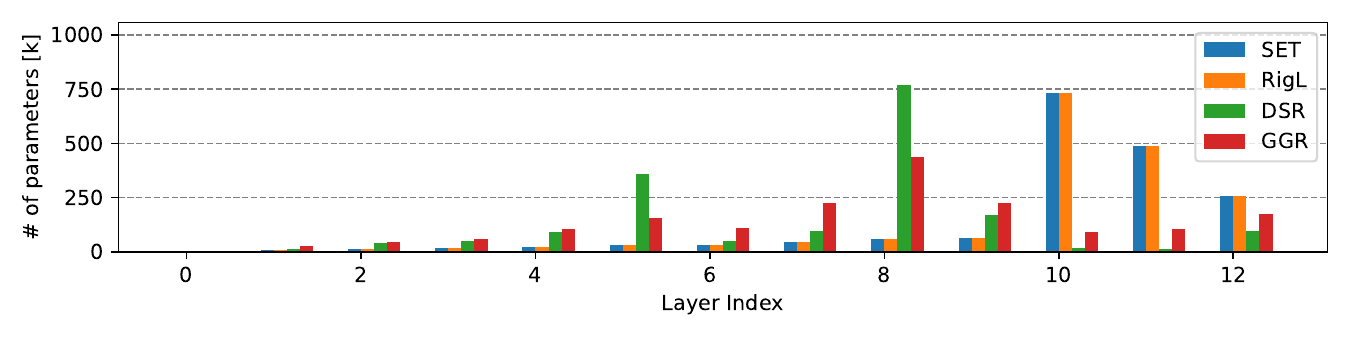}
    \vspace{-20pt}
    \caption{Layer density and number of active parameters per layer in VGG16 trained on Tiny-ImageNet with 97\% sparsity using \erdosrenyi-Kernel initialization.
    }
    \label{fig:layer_densities_parameters_vgg16}
\end{figure*}

We plot the final densities and number of parameters of each layer in VGG16 trained with 97\% sparsity and \erdosrenyi-Kernel initialization in Figure~\ref{fig:layer_densities_parameters_vgg16}.
These results give us insights into the differences between DSR and \ours distribution.
DSR tends to target more extreme sparsity levels in comparisons to \ours, due to their heuristic which punishes sparser layers by inserting less weights into them.
The tenth and eleventh layers are large linear layers that should be sparser than the rest of the layers.
While \ours increases the sparsity for these layers significantly, DSR drains these layers even more, making them one of the layers with the lowest number of weights while being by far the largest parameter tensors in the dense case.
Overall, \ours seems to balance the number of parameters per layer more evenly than DSR.
\erdosrenyi-Kernel excels at inserting more weights into the smaller layers, but still assigns much more weights into the large layers 10 and 11 than both dynamic approaches.
In a neural network with such unbalanced parameter sizes, \erdosrenyi-Kernels seems unable to get the most out of the given parameter budget.

\begin{figure}[ptb]
\centering
\begin{subfigure}[b]{0.248\linewidth}
  \includegraphics[width=.95\textwidth]{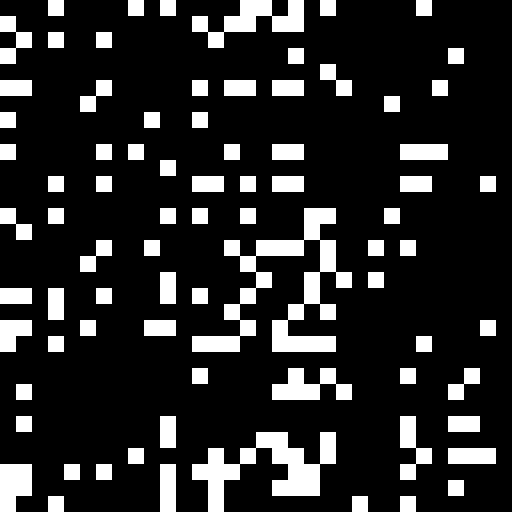}
  \caption{SET}
\end{subfigure}%
\begin{subfigure}[b]{0.248\linewidth}
  \includegraphics[width=.95\textwidth]{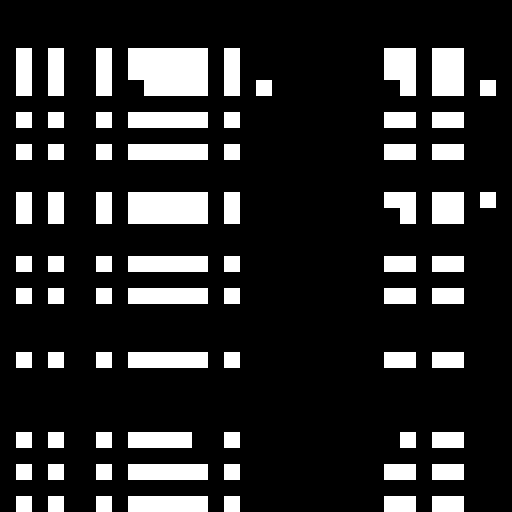}
  \caption{RigL}
\end{subfigure}%
\begin{subfigure}[b]{0.248\linewidth}
  \includegraphics[width=.95\textwidth]{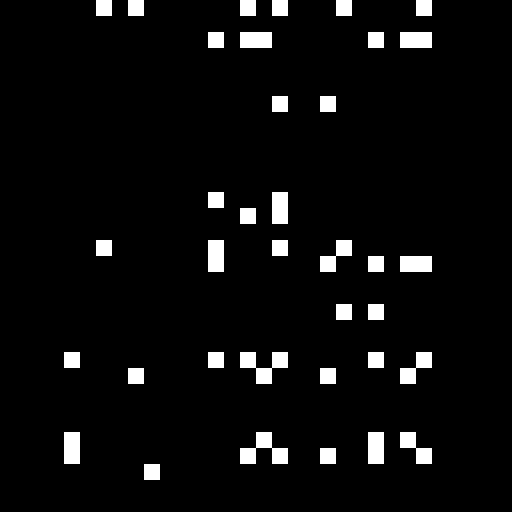}
  \caption{DSR}
\end{subfigure}%
\begin{subfigure}[b]{0.248\linewidth}
  \includegraphics[width=.95\textwidth]{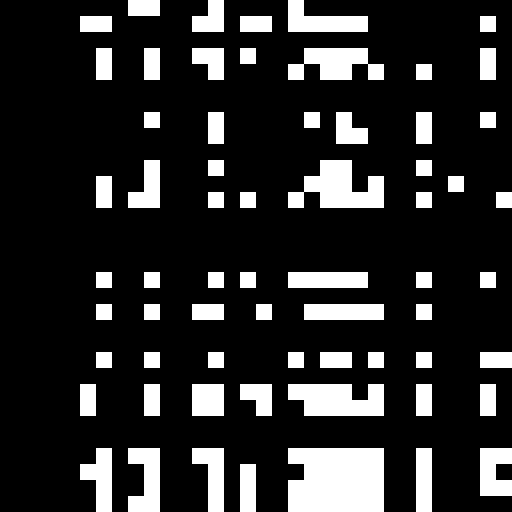}
  \caption{\ours}
\end{subfigure}%

\caption{
Parameter masks from the third layer (first convolutional layer in stage 1) in MobileNetv2 trained on CIFAR-100 with 97\% sparsity and \erdosrenyi-Kernel initialization. White pixels indicate active weights.
}
\label{fig:parameter_distribution}
\end{figure}

In Figure~\ref{fig:parameter_distribution}, we illustrate one representative example of the final parameter mask each evaluated approach converges to.
SET, using random insertion, ends up with a more uniformly distributed mask compared to RigL, yet with some rows and columns having a higher density than others.
RigL focuses much more on few rows and columns that get densely populated while the rest stays nearly empty.
DSR, like SET, uses random insertion and therefore creates a more randomly populated mask compared to the structured mask from RigL.
The parameter mask formed by \ours displays what we expect from using a combination of random and gradient-based insertion.
There are clear patterns in the mask with some densely populated rows and columns, but it is less structured than in the case of RigL.
\section{Discussion}
\ours shows promising results, achieving the highest accuracy at extreme sparsity levels.
Our approach is less sensitive to initialization than all existing approaches - showing that it has a great potential for networks that are more difficult to balance optimally.

Our evaluations show that there is no single solution that always achieves the best results in all scenarios - as is the case with optimizers, learning rate schedulers and hyper-parameters like batch size and epochs. 
However, when looking at the average performance over all four dataset and model configurations, \ours performs best overall. 
For the sake of transparency and representative comparability, we avoided tuning parameters in a way that benefits our approach most but tuned them only for the dense reference.
While sparse approaches might achieve slightly better results than in our evaluations when performing a grid-search over hyper-parameters for each approach, we chose to use one set of hyper-parameters that is shared over all approaches.
This reduces the cost of training, while also being more representative in our opinion, as the results show how each approach performs as a drop-in replacement for the dense network.

Combining gradient-based and random weight insertion is crucial for an overall consistent performance - at least when combined with dynamic density balancing.
Using only gradient-based or random insertion can lead to better results in some scenarios, but also to much worse results in others.
We think that using one approach that always performs close to the best is important for establishing dynamic sparse training as the next step forward to more efficient training and inference. 

\ours is good at recovering from bad initialization. 
At the same time, it does not make good initialization obsolete.
Balancing weight densities with \erdosrenyi-Kernel leads to better results in almost all cases - and it allows for training networks at extreme sparsities which otherwise fail entirely.

The results in Table~\ref{tab:empty_neurons_inputs} show that \ours is able to train the model to the highest accuracy while using the lowest number of neurons and inputs.
This might be counter-intuitive at first, given that we can shrink VGG16 to a small, dense model which achieves better results than other methods with larger, sparser models.
We assume that the benefit of large, sparse models mainly applies to the training process in which the larger set of subnetwork candidates is advantageous over the small, dense network. 
Yet, not all neurons are required in the final model - especially when the model is overparameterized as is the case in the later layers of VGG16.
In contrast, no method was able to shrink ResNet-18 significantly, since over 99\% of neurons and inputs were used in the final, converged network at the same sparsity level of 97\%.
Compared to VGG16, however, ResNet-18 uses much less parameters and does not feature excessively large linear layers.
The fact that no approach was able to shrink the model might indicate that it is better designed or better suited for the dataset it is trained on.
\section{Conclusion}
In this work, we implemented and evaluated a dynamic density adjusting approach for dynamic sparse training.
We determine the number of weights to be inserted in each layer using the top-k gradients of all layers.
\ours achieves state-of-the-art performance in all settings - with the greatest leap at high sparsity levels.
Using \ours, extremely sparse neural networks can be trained more reliably than with previous work - even when the parameters are badly distributed at the start.
Furthermore, we showed that using a combination of gradient-based and random insertion helps to improve consistency in training neural networks at different sparsity levels.
While our method can benefit from good initialization, it still achieves good results when initialized badly, which is a great advantage when training complex neural networks with extreme sparsity.

\bibliography{references}
\bibliographystyle{icml2022}

%%%%%%%%%%%%%%%%%%%%%%%%%%%%%%%%%%%%%%%%%%%%%%%%%%%%%%%%%%%%%%%%%%%%%%%%%%%%%%%
%%%%%%%%%%%%%%%%%%%%%%%%%%%%%%%%%%%%%%%%%%%%%%%%%%%%%%%%%%%%%%%%%%%%%%%%%%%%%%%
% APPENDIX
%%%%%%%%%%%%%%%%%%%%%%%%%%%%%%%%%%%%%%%%%%%%%%%%%%%%%%%%%%%%%%%%%%%%%%%%%%%%%%%
%%%%%%%%%%%%%%%%%%%%%%%%%%%%%%%%%%%%%%%%%%%%%%%%%%%%%%%%%%%%%%%%%%%%%%%%%%%%%%%
% \newpage
% \appendix
% \onecolumn
% \section{You \emph{can} have an appendix here.}

% You can have as much text here as you want. The main body must be at most $8$ pages long.
% For the final version, one more page can be added.
% If you want, you can use an appendix like this one, even using the one-column format.
%%%%%%%%%%%%%%%%%%%%%%%%%%%%%%%%%%%%%%%%%%%%%%%%%%%%%%%%%%%%%%%%%%%%%%%%%%%%%%%
%%%%%%%%%%%%%%%%%%%%%%%%%%%%%%%%%%%%%%%%%%%%%%%%%%%%%%%%%%%%%%%%%%%%%%%%%%%%%%%

\end{document}